# HOW TO DO THINGS WITH PROMPTS

**Kristina Šekrst, Virna Karlić**

When users address large language models, they produce directive speech acts whose pragmatic features differ from those of both everyday conversation and traditional human-computer interaction, and these features change as users gain familiarity with the systems they address. This paper applies speech act and politeness theory to a corpus-pragmatic analysis of 2,000 English-language prompts drawn from publicly shared ChatGPT conversations, 1,000 from 2023 and 1,000 from 2025, using the ShareChat dataset. Each prompt is annotated for illocutionary force, directness, propositional content, and the presence of politeness markers, and the distribution of these features is compared across the two sampling years. The results show a consistent movement toward indirect, implicit, and fragmentary realizations of directive force, accompanied by a decline in politeness marking. The largest single change, a shift of 14.9 percentage points, occurs in propositional content, where explicit specification of the requested action gives way to implicit reliance on the system's inferential capacity, suggesting that users have updated their model of what the system can recover from reduced input, treating it as a competent implicature resolver. Rather than asking whether LLMs "really" understand language, we should ask: what kind of language have we created in learning to speak to them?

## 1. Prompts as performative speech acts

With the growing use of generative AI chatbots, specific pragmatic features of interaction between human (expert and non-expert) users and AI systems have begun to emerge. That is, generative AI chatbots have spread quickly enough that the way people speak to them is acquiring its own kind of pragmatic turn. As Holtel (2026, p. 118) points out, “The new interface paradigm, exemplified by AI-driven chatbots such as those developed by OpenAI (2022), enables forms of communication that mirror techniques of human dialogue more than ever before.” Our starting intuition came from watching our own habits. Namely, as ordinary users, we noticed our prompts growing terse and imperative, even a little draconian, commands tossed at something built to comply. However, as technical users, we watched prompt engineering harden into a craft, one that rewards detail and precision as the task grows more demanding. This paper follows the first group, the general audience that treats large language models as conversational partners, because that kind of exchange runs on pragmatic cues interesting to track as they change over time.

This development makes it possible to position prompt engineering within the broader framework of linguistics, particularly discourse analysis and pragmatics. In this context, prompt crafting, traditionally treated as a “technical exercise involving precise command inputs”, may be reconceptualized as a “dialogical and intentional communicative practice that derives from human language” (Holtel 2026, p. 118). Following this approach, the present paper applies the fundamental principles of Austin’s and Searle’s speech act theory, as well as Brown and Levinson’s (1987) politeness theory, in order to examine the pragmatic features of prompts as a specific type of performative speech act addressed by human users to AI chatbots as their “interlocutors”.

The concept of a speech act is defined as “actions performed via utterances” (Yule 1996, p. 47) and holds that language use constitutes a form of human action carried out through speech acts such as making statements, giving orders, and asking questions. The founder of speech act theory was John Austin, who, together with John Searle, laid the groundwork for one of the most influential pragmatic theories of language. In his seminal work *How to Do Things with Words*, Austin (1962) distinguishes between *constative* utterances, through which the speaker makes claims that can be evaluated as true or false, and *performative* utterances, through which the speaker performs an act that can be evaluated as successful (*felicitous*) or unsuccessful (*infelicitous*). In the context of human-AI chatbot interactions, prompts from human users are generally interpreted as performative utterances in which users seek a response from the system.

Today, numerous classifications of speech acts exist according to different criteria, but the most influential was proposed by Searle, who systematically elaborated Austin’s theory in *Speech Acts: An Essay in the Philosophy of Language* (1969) and his later works. According to Searle’s classification (1975), speech acts are divided into five groups: (1) *representatives* (assertives), through which speakers make statements that can be evaluated as true or false; (2) *directives*, through which speakers attempt to influence the listener’s actions or behavior (e.g., commands, requests, pleas, advice); (3) *commissives*, through which speakers commit themselves to future action (e.g., promises, threats, offers); (4) *expressives*, through which speakers express feelings or attitudes (e.g., congratulations, apologies, accusations); (5) *declarations*, through which, with appropriate institutional authority, the state of affairs in extralinguistic reality is changed (e.g., declarations of war, nominations, excommunications) (briefly adapted from Karlić and Bago, 2021, p. 38–39).

In user-chatbot communication, prompts produced by human users can, in most cases, be classified as directives, namely *task-oriented prompts* (Example 1a) and *information-seeking prompts* (Example 1b), through which users ask the system to perform a task or answer a question. Although every prompt is inherently designed to elicit a response from the system and, therefore, in this sense, displays the basic characteristics of directives, some prompts serve exclusively to simulate interpersonal communication (e.g., expressions of gratitude, praise, greetings). Such prompts may therefore be characterized as *phatic prompts*, whose pragmatic function corresponds to Searle's category of expressive speech acts (Example 1c).

**Example 1**[1]

(a) Look up relevant betting markets including manifold and make predictions for them
(b) Does <REDACTED> have husband or wife?
(c) I'll see you later

For each type and subtype of speech act, Searle (1969) defines the conditions necessary for its successful performance and further analyzes speech acts in terms of two semantic components: (1) *propositional content* and (2) *illocutionary force*. These components may be expressed explicitly or implicitly at the level of propositional content, and directly or indirectly at the level of illocutionary force. In the context of human-AI chatbot interaction, Searle's classification may be applied to prompts that display the characteristics of directive speech acts in the following manner:

**Example 2**

**Prompt A:** give the movement/function of the following muscles in short sentences each

| **Type of speech act** | directive (request/instruction) |
|---|---|
| **Illocutionary force** | explicit: imperative verb *give* |
| **Propositional content** | explicit: description of muscle movement/function |

**Prompt B:** All the code needed for my project transcendence cybernetic ascension

| **Type of speech act** | directive (request/instruction) |
|---|---|
| **Illocutionary force** | implicit: incomplete nominal construction |
| **Propositional content** | explicit: generation of all code needed for the project |

Searle further emphasizes that every utterance possesses a specific illocutionary purpose, that is, a communicative goal. For example, all directive speech acts share the same illocutionary purpose – influencing the behavior of the addressee, although they may differ in terms of illocutionary force (e.g., a request and a command). According to the politeness theory of Penelope Brown and Stephen Levinson (1987), directive speech acts are inherently intrusive because their purpose is to induce the interlocutor to perform a particular action; consequently, they may function as face-threatening acts.

[1] All examples are drawn from the corpus analyzed later in the paper (see the methodology section). The concrete examples in the paper were sampled from the smaller corpus of 100 examples, with which we started the initial analysis.

> The degree of illocutionary force depends on the extent to which the speaker's intention, expressed explicitly or implicitly through a directive, is binding upon the addressee. If the speaker strongly insists on the realization of the requested action, the illocutionary force becomes stronger. Conversely, the less obligatory the directive is for the addressee, the weaker or more mitigated its illocutionary force becomes. (Karlić & Bago, 2021, p. 200)

In interpersonal communication, face-threatening acts such as directive speech acts with strong illocutionary force may be mitigated through face-saving politeness strategies, which can be verbal (e.g., lexical devices, grammatical constructions, indirectness/implicitness) and non-verbal (e.g., intonation, facial expressions) (Karlić & Bago, 2021). For example, the illocutionary force of the direct directive speech act *Turn the music down* may be softened using the lexical marker *please* (command → request/plea), conditional and/or interrogative constructions (*Could you turn the music down?*), as well as by implying the propositional content (*The music is too loud*).

In conversational AI exchanges, one of the primary mechanisms for creating human-like communication is the (mutual) use of politeness strategies, such as phatic prompts (Example 1c) and devices used to mitigate the illocutionary force or to perform other interactional functions, such as greetings, expressions of gratitude, and praise (Example 3).

**Example 3**

<u>Great work thank you</u>, great length! One reminder: <u>please</u> do not transition to future topics

On the other hand, in interpersonal communication, the mitigation of face-threatening acts represents an essential strategy for maintaining cooperative social interaction. AI chatbots do not possess social "face" needs that could be threatened in a manner that would negatively affect their cooperative behavior. From a purely functional perspective, the use of politeness strategies is, therefore, not necessary for the successful performance of the communicative act.[2] Moreover, such strategies may be considered costly and less energy-efficient, as they require a greater number of words and consequently higher energy consumption. Reflecting on this phenomenon, OpenAI CEO Sam Altman remarked that “Politeness in ChatGPT prompts like ‘please’ and ‘thank you’ costs OpenAI tens of millions annually.” He nevertheless described this expense as “money well spent”, since users appear to prefer such interactional behavior – whether out of habit or a form of “futuristic caution” (Hanif 2025).

The pragmatic features of prompts have consequently become the subject of numerous discussions in the media and on social networks, typically involving two opposing positions: users inclined to simulate polite interpersonal conversation and those who consider such behavior unnecessary. As a result of these discussions and users gradually becoming accustomed to interacting with AI chatbots, the pragmatic features of interaction between human users and AI systems seem to have changed since the early stages of their widespread use. The assumption that this interaction has evolved over time represents the main hypothesis explored in the remainder of the paper.

## 2. Corpus-Pragmatic Analysis

### 2.1 Methodology

Drawing on the theoretical framework outlined in the previous section, we compiled a pragmatically annotated corpus of prompts from non-professional human users who interacted with ChatGPT in 2023 and 2025 to examine the characteristics of these prompts as speech acts. The two time points were chosen deliberately because they bracket a period when conversational AI moved from novelty to routine for many users, making any drift in how people phrase their requests visible over the interval. By non-professional, we mean users with no formal training in prompt engineering and no technical stake in optimizing model output (as we would expect the technical users to drift more towards details and accuracy, especially regarding few-shot examples),[3] the population whose linguistic habits carry over most directly from ordinary human conversation. Each prompt was coded for illocutionary force, directness, and the presence or absence of politeness markers, which lets us track whether the pragmatic shape of a request hardened, softened, or held steady over the two years.

[2] However, some studies (e.g., Yin et al., 2024) suggest that violations of politeness principles may affect LLM performance: "The study notes that using impolite prompts can result in the low performance of LLMs, which may lead to increased bias, incorrect answers, or refusal of answers. However, highly respectful prompts do not always lead to better results. In most conditions, moderate politeness is better, but the standard of moderation varies by languages and LLMs."

### 2.1.1 Research questions and the hypothesis

The corpus was designed to address the following research questions (Q) and test the proposed hypothesis (H), as seen in Table 1:

**Table 1: Research questions and hypothesis details.**

| | |
|---|---|
| **Q1** | What is the distribution of directive and phatic prompts? |
| **Q2** | What is the distribution of prompts displaying the characteristics of direct and indirect directive speech acts? |
| **Q3** | What is the distribution of different subtypes of direct and indirect directive speech acts? |
| **Q4** | What is the distribution of prompts displaying directive speech acts with explicitly and implicitly expressed propositional content? |
| **Q5** | What is the distribution of prompts with and without the use of politeness strategies? |
| **H1** | The pragmatic features of prompts produced by non-professional human users changed between 2023 and 2025. |

### 2.1.2 Corpus Compilation and Annotation

The corpus was compiled from ShareChat[4] (Yan et al. 2026), a large-scale collection of 142,808 conversations (660,293 turns) sourced directly from publicly shared URLs across five major chatbot platforms: ChatGPT, Perplexity, Grok, Gemini, and Claude. The dataset spans 101 languages and covers the period from April 2023 to October 2025, making it the first cross-platform corpus to preserve native interface affordances such as citations, thinking traces, and code artifacts rather than stripping interactions down to plain text as prior datasets do. The ChatGPT subset, which we draw on here, comprises 102,740 conversations and 542,148 turns, with an average of 5.28 turns per conversation. Conversations entered the dataset through the platforms' own share function, which means the data reflects interactions that users elected to make public rather than conversations logged through a research gateway or proxy interface, a distinction that reduces the observer bias introduced when participants know they are being monitored. Because shared conversations originate with ordinary users rather than professional prompt engineers or annotators, the material reflects the prompting practices of non-professional users – the population the present study addresses.

[3] A technical user, by contrast, tends to drift toward detail and accuracy, and the clearest marker of this is the shift from zero-shot to few-shot prompting. A *zero-shot prompt* states the task and expects the model to perform it without worked examples, as in "Classify this review as positive or negative: The plot dragged but the acting saved it." A *few-shot prompt* supplies several labeled demonstrations before the target case, so the request becomes "Review: Stunning visuals, weak script. Sentiment: mixed. Review: Boring from start to finish. Sentiment: negative. Review: The plot dragged but the acting saved it. Sentiment:" and the model completes the pattern. The engineering literature treats this scaffolding as a reliability technique, and related practices such as chain-of-thought prompting, where the user appends an instruction like "Let us reason step by step", push further in the same direction (Skansi & Šekrst, 2026). None of this machinery appears in the speech of users who simply ask the model for what they want. The non-professional prompt stays close to an ordinary directive, and that is precisely why its pragmatic features repay analysis.
[4] The dataset is publicly available on Hugging Face under a dedicated license agreement: https://huggingface.co/datasets/tucnguyen/ShareChat

We chose GPT because the launch of ChatGPT – the accessible interface for the GPT model, including various tools and file upload – was the first to introduce LLM-based AI to a wider audience, and that early reach lets us track the difference in prompting for the general public across the whole of 2023 and again in 2025. For comparison, Anthropic's competitor model, Claude, launched in March 2023 and spread more gradually, so its early usage data would not give us the consistent, high-volume signal the analysis requires.

From this subset, we automatically extracted the English-language conversations using Python's *langdetect* library, followed by an additional pass with Claude Opus 4.7 and a human overview after the automatic task. We retained the whole conversation as the unit of analysis, including various turns, since we wanted to see how the users refer to the conversational context as well. Of the 1000 prompts sampled for each year, 462 (2023) and 417 (2025) are conversation-initial turns, and the remainder are continuation turns that rely on prior conversational context.

We first sampled 100 random prompts from the 2023 portion of the data and 100 from the 2025 portion that were annotated manually, and then gave two LLMs (GPT 5.5 and Claude Opus 4.7) another 100 prompts to annotate. The accuracy compared with 4 human annotators was comparable, so we sampled 1,000 prompts from the 2023 portion of the data and 1,000 from the 2025 portion, yielding a balanced corpus of 2,000 prompts that permits comparison across the two sampling years. The two years were selected to represent an early stage of widespread public use of ChatGPT and a later stage at which users had become accustomed, both socially and linguistically, to interacting with the system.

As mentioned above, the annotation proceeded in two stages: (1) We annotated a calibration set of 100 prompts manually, applying the schema described below (See Appendix A) in order to establish consistent criteria for each pragmatic feature. Claude Opus (version 4.7) was then prompted to annotate the entire corpus using the same schema, while GPT 5.5 was used for a second pass. (2) Each automatically assigned label was subsequently checked and, where necessary, corrected by the authors against the criteria fixed during the manual calibration stage, so that the final annotations reflect human judgment rather than the model's output alone.

The annotation of corpus examples involved the analysis of the pragmatic features presented in the following Table 2 (see also Appendix A):

**Table 2: pragmatic classification of features.**

| 1. PROMPT TYPE | | |
|---|---|---|
| **Directive** | Requesting information or task execution. | *Translate the song.* |
| **Phatic** | Serving primarily interpersonal/interactional functions | *Hello!* |
| **2. DIRECTIVES: DIRECT ILLOCUTIONARY FORCE** | | |
| **Imperative** | Imperative verb requesting task performance. | *Translate the song.* |
| **Performative** | Performative verb requesting task performance. | *I ask you to translate the song.* |
| **Info-seeking interrogative** | Information-seeking interrogative requesting factual, explanatory, or interpretive information. | *Who wrote this song* |
| **3. DIRECTIVES: INDIRECT ILLOCUTIONARY FORCE** | | |
| **Assertive** | Directive realized through an assertive form. | *I need this song translated.* |
| **Task-seeking Interrogative** | Directive realized through an interrogative form. | *Can you translate this song?* |
| **Incomplete** | Directive realized through a sentence fragment or incomplete construction. | *translate song* |
| **Raw input** | Raw textual or symbolic material submitted without any user’s instruction. | [song lyrics] |

| 4. DIRECTIVES: PROPOSITIONAL CONTENT | | |
|---|---|---|
| **Explicit** | The requested action is explicitly named. | *Translate.* |
| **Implicit** | The requested action is only implied. | *Here's the song.* |
| **5. POLITENESS/INTERACTIONAL MARKERS** | | |
| **Yes** | Presence of politeness or interactional markers. | *Translate, please.* |
| **No** | No politeness or interactional markers. | *Translate the song.* |

## 2.2 Results

The annotated corpus comprises 1,000 English-language prompts for each sampling year (2023 and 2025), produced by non-professional users in shared ChatGPT conversations. The results are reported below in the order of the research questions set out in Section 2.1.1, with each prompt feature quantified for both years and the change expressed in percentage points (pp). The sampled corpus information is available in Appendix B.

### 2.2.1 Directive and phatic prompts (Q1)

Directive prompts dominate the corpus in both years and account for the overwhelming majority of the material. Phatic prompts, whose function corresponds to Searle's category of expressive speech acts (Example 1c), remain marginal and decline slightly over the period. Their share drops from 2.3% in 2023 to 1.6% in 2025, indicating that the corpus is, in pragmatic terms, almost entirely composed of task-oriented and information-seeking exchanges rather than interpersonal ones (see Table 3).

**Table 3: Distribution of directive and phatic prompts.**

| Prompt type | 2023 | 2025 | Change |
|---|---|---|---|
| **Directive** | 977 (97.7%) | 984 (98.4%) | +0.7pp |
| **Phatic** | 23 (2.3%) | 16 (1.6%) | -0.7pp |

The absolute size of the phatic category is small in both years, which limits the inferences that can be drawn from the year-on-year difference alone. The character of the prompts coded as phatic is nonetheless informative. Namely, the category includes conventional greetings, expressions of gratitude, sign-offs, and short reactive turns of the type illustrated by Example 1c, and in both samples, these tend to occur as part of an exchange that is otherwise directive rather than as the sole content of a session. The residual phatic material reflects the habitual transfer of interpersonal conventions into interactions with a chatbot (cf. Hanif, 2025), rather than a sustained attempt to maintain an interactional relationship with the system.

The predominance of directive prompts is also unsurprising: users address the system in order to obtain something from it, and the directive is the speech-act type through which that function is performed. However, the relevant observation is the direction of the small change between the two years: the corpus becomes marginally more directive and marginally less phatic over the period, in a movement consistent with the larger shifts in illocutionary force, propositional content, and politeness marking documented in the sections that follow.

### 2.2.2 Direct and indirect directives (Q2)

Among directive prompts, direct realizations predominate in both years, although their share contracts over the period. Direct directives fall from 65.7% in 2023 to 58.3% in 2025, while indirect realizations rise correspondingly from 34.3% to 41.7%, a shift of 7.4 percentage points, visible in Table 4. The movement suggests that users increasingly rely on forms whose directive force is conveyed through means other than an explicit imperative or information-seeking question.

**Table 4: Distribution of direct and indirect directives.**

| Illocutionary force | 2023 | 2025 | Change |
|---|---|---|---|
| Direct | 642 (65.7%) | 574 (58.3%) | -7.4pp |
| Indirect | 335 (34.3%) | 410 (41.7%) | +7.4pp |

In interpersonal communication, the choice between the two is conditioned by considerations of politeness and face, since indirectness mitigates the imposition that directives carry by their nature, but the cost of indirectness, on the other hand, is a measure of interpretive effort transferred to the addressee, who must recover the requested action from a form that does not name it. The change documented in Table 4 has a specific reading in the human-AI setting. The mitigating function of indirectness, central to its interpersonal use, has no counterpart in chatbot exchanges, since the system has no face to save, and the growth of indirect realizations cannot therefore be attributed to politeness. The remaining motivation for indirectness is the one identified above: the transfer of interpretive work from the speaker to the addressee.

The higher rate of indirect forms in 2025 is consistent with users relying more heavily on the system to recover the intended directive from material that does not state it explicitly. The corresponding decline in direct realizations is the visible side of this reliance, and the 7.4-percentage-point aggregate shift sets the frame for the more pronounced internal redistributions documented in the following section, where the indirect subtype distribution indicates which forms absorb the resulting share. The growth of indirect realizations brings the corpus closer to the conventions of interpersonal communication in terms of form, since indirectness and implicature presuppose an inferentially capable addressee, and the 2025 distribution reflects a user population that has updated its assessment of what the system can recover, as users come to treat the model as a competent resolver of implicature rather than a literal-minded executor of explicit instructions.

### 2.2.3 Subtypes of direct and indirect directives (Q3)

Within the direct directives, imperatives and information-seeking interrogatives occur at comparable frequencies, together accounting for the bulk of the direct material. The direct subtype codes in Table 5 are non-exclusive: a small but growing class of direct prompts combines an information-seeking interrogative with an imperative that constrains the system's response, as in "Does measurement require two? Limit your answer to yes or no" or "What ideology would you most closely relate to the gesture. Answer in one word." Seven such hybrid prompts occur in 2023, and eleven in 2025 – each is counted under both the imperative and the information-seeking interrogative subtypes. The subtype totals, therefore, overshoot the direct total reported in Table 4 by 7 and 11, respectively. Collapsing these prompts into a single subtype would obscure a pragmatic feature of the later register, namely the practice of pairing a question with an output-formatting directive, and the non-exclusive coding preserves the pattern for analysis. The indirect subtypes in Table 6 are coded as mutually exclusive by contrast, since their hybrid cases (two prompts in the 2025 sample) are realizations of a single request through two competing forms rather than two distinct directives acting on different aspects of the response.

**Table 5: Distribution of direct directive subtypes as a share of all directives.**

| Direct subtype | 2023 | 2025 | Change |
|---|---|---|---|
| **Imperative** | 333 (34.1%) | 276 (28.0%) | -6.0pp |
| **Performative** | 0 | 0 | 0pp |
| **Info-seeking interrogative** | 316 (32.3%) | 309 (31.4%) | -0.9pp |

Imperative prompts decline from 34.1% to 28.0%, a drop of 6.0 percentage points, while information-seeking interrogatives remain effectively stable at roughly one third of all directives. Performative formulations of the type illustrated by the codebook example “I ask you to translate the song” are absent in both years, confirming that explicit metapragmatic framing of the request plays no role in this register.

The two surviving direct subtypes correspond to the two structurally simplest realizations of a directive in English: the imperative names the action and requests its performance, and the information-seeking interrogative names the gap in knowledge and requests its closure. Each is, in Searle's terms, a direct realization in which sentence form and illocutionary force are aligned, and each requires the addressee to recover nothing beyond what the surface form already supplies. The absence of performative formulations in both samples is consistent with this picture. An explicit performative, such as "I ask you to translate", performs the directive through a verb that names the act of requesting, and the construction is more heavily pragmatic than the equivalent imperative or interrogative, without delivering any additional information. In interpersonal use, it persists where institutional or interpersonal stakes make the act of requesting itself worth marking; when the human-AI exchange offers no such stakes, the result is a register in which the performative slot is effectively empty.

The decline in imperatives accounts for most of the contraction in direct realization documented in Table 5. The 6.0-percentage-point drop in imperatives is matched, in the opposite direction, by the growth of indirect realizations analyzed below in this section, and the residual share of information-seeking interrogatives is broadly preserved. Imperatives are therefore the direct subtype whose role declines over time, while questions seeking factual or interpretive information continue to be addressed to the system at roughly the same rate. The reason for these changes is that imperatives compete with the indirect forms that absorb the share, since both are mechanisms for requesting a task, and the alternatives available to a user requesting a task have proliferated. Information-seeking interrogatives have no comparable competitor, since their function – the closure of an epistemic gap – is performed most naturally by a direct question and has no fragmentary or assertive analog in the indirect repertoire.

To illustrate further, a request that in 2023 would have been issued as the full imperative "translate this to English" can in 2025 be reduced to the prepositional fragment "to English" without loss of communicative function, since the operand alone suffices once the user can rely on the system to recover the verb from the preceding turn or from conversational context. The fragment "to English" is, in the codebook's terms, an incomplete-sentence realization of the same directive that the full imperative performs, and it competes with the imperative for the same communicative slot. No equivalent reduction is available for the information-seeking interrogative. However, the question "how do you say this in French" cannot be reduced to "in French" without losing the epistemic-gap signal that distinguishes a question from a translation request, since the interrogative form itself carries information about function that a bare nominal does not preserve.

**Table 6: Distribution of indirect directive subtypes.**

| Indirect subtype | 2023 (n=335) | 2025 (n=410) | Change |
|---|---|---|---|
| **Assertive** | 172 (51.3%) | 195 (47.6%) | -3.8pp |
| **Task-seeking interrogative** | 83 (24.8%) | 51 (12.4%) | -12.3pp |
| **Incomplete** | 42 (12.5%) | 89 (21.7%) | +9.2pp |
| **Raw input** | 38 (11.3%) | 75 (18.3%) | +7.0pp |

The indirect directives display a more pronounced reorganization, as seen in Table 6. Assertive realizations of the type “I need this song translated” remain the most frequent subtype throughout, though their share falls from 51.3% to 47.6%. Task-seeking interrogatives, which carry directive force through interrogative form, decline sharply from 24.8% to 12.4%, a fall of 12.3 percentage points. The reduction is offset by growth in the two most reduced forms. Incomplete constructions, in which the request is realized through a sentence fragment, rise from 12.5% to 21.7%, and raw input prompts, consisting of textual or symbolic material submitted without any framing instruction, rise from 11.3% to 18.3%. Taken together, fragmentary and unframed realizations account for 23.8% of indirect directives in 2023 and 40.0% in 2025, which represents the most substantial internal redistribution observed in the corpus.

### 2.2.4 Propositional content (Q4)

Among task-oriented directives, the balance between explicit and implicit propositional content reverses over the period.[5] In 2023, the requested action is explicitly named in 62.9% of cases and only implied in 37.1%. By 2025, the proportions have crossed, with explicit content falling to 48.0% and implicit content rising to 52.0%, a shift of 14.9 percentage points and *the largest single change recorded in the analysis* (see Table 7). The pattern indicates that users increasingly leave the requested action to be inferred from the surrounding context rather than stating it directly, as illustrated by Prompt B in Example 2.

**Table 7: Distribution of explicit and implicit propositional content.**

| Propositional content | 2023 | 2025 | Change |
|---|---|---|---|
| **Explicit** | 416 (62.9%) | 324 (48.0%) | -14.9pp |
| **Implicit** | 245 (37.1%) | 351 (52.0%) | +14.9pp |

The annotation picks out whether the prompt names the action verb. A prompt with explicit content states what the system is to do, as in "translate the text below" or "summarize this article". A prompt with implicit content does not name the action but expects the system to recover it from context: "here is the song" (translation implied by the preceding turn), "below is my essay" (editing or commenting implied by the genre), a pasted code snippet submitted without instruction (explanation or completion implied by the conventions of programming-related interaction).

[5] The base population is 661 prompts in 2023 and 675 in 2025, since propositional content is annotated only for task-oriented directives. Information-seeking interrogatives are excluded, on the grounds that the propositional content of a question is the question itself rather than a separately specifiable action, so the explicit-or-implicit distinction does not apply in the same sense.

The change from Table 7 is partially driven by the inclusion of continuation turns in the corpus. When the analysis is restricted to conversation-initial prompts only, the explicit-to-implicit ratio moves from 67.6% explicit in 2023 to 55.4% explicit in 2025, which is a substantial shift of 12.2 percentage points, but one that does not cross the 50% threshold (see Table 8). Continuation turns, which inherently rely on context established in prior exchanges, are more likely to leave the requested action implicit (e.g., "Continue", "Do the same for the rest", "Try again"). The proportion of continuation turns in the corpus is itself high (53.8% in 2023 and 58.3% in 2025), and this structural feature of multi-turn conversation amplifies the aggregate tendency toward implicitness. This finding, therefore, reflects both a change in how users formulate individual prompts and a tendency for shared conversations to extend over multiple turns, each of which compounds the reliance on conversational context.

**Table 8: Explicit propositional content by prompt position.**

| Prompt position | 2023 | 2025 | Change |
|---|---|---|---|
| **All task-oriented directives** | 416/661 (62.9%) | 324/675 (48.0%) | -14.9pp |
| **First prompts only** | 200/296 (67.6%) | 139/251 (55.4%) | -12.2pp |
| **Continuation turns** | 216/365 (59.2%) | 185/424 (43.6%) | -15.5pp |

Propositional content and illocutionary force are formally independent in Searle's framework, but the corpus shows they interlock. Users do less of the work of specifying what the system is to do: the fact that this dimension records the largest movement in the corpus, well beyond any feature of illocutionary form, suggests that the principal site of change is not how users frame their requests but how much they specify in the first place. The syntactic change toward *indirectness*, *fragmentation*, and *reduced politeness* follows from a population that has decided to delegate more of the underlying content to the system, and the size of the propositional-content effect is the clearest indication that the delegation is the central phenomenon rather than a side effect.[6]

[6] One caveat is worth mentioning: models reason better and the context window – the amount of data they can "look at" at the same time has increased from 2023 to 2025 (cf. Skansi & Šekrst, 2026), but that change is not as significant for the first message that is the part of the dataset, where the context

### 2.2.5 Politeness and interactional markers (Q5)

Politeness and interactional markers are present in a minority of directives in both years and become less frequent over the period: their share falls from 16.2% in 2023 to 10.9% in 2025, a decline of 5.3 percentage points, as seen in Table 9. The result is consistent with the interpretation that users treat the system as an instrument for task completion rather than as a social interlocutor whose face needs to be mitigated, and it aligns with the absence of performative framing and the marginal status of phatic prompts already noted.

**Table 9: Distribution of politeness and interactional markers.**

| Politeness markers | 2023 | 2025 | Change |
|---|---|---|---|
| **Present** | 158 (16.2%) | 107 (10.9%) | -5.3pp |
| **Absent** | 819 (83.8%) | 877 (89.1%) | +5.3pp |

The pragmatic reading is again in light of the framework set out in Section 1: politeness strategies arise in interpersonal communication as a response to the face-threatening character of directives, and the chatbot exchange offers no addressee with a face that can be threatened in the relevant sense, so the strategies have no functional work to do beyond whatever the user transfers from interpersonal habit. However, the residual category is informative in its own right, since even in 2025, roughly one prompt in ten still carries some form of politeness or interactional marking, which is too large to attribute to incidental traces and suggests that the futuristic-caution motivation Altman gestures at (Hanif, 2025) operates as a stable minority disposition rather than as a behavior that the whole population is shedding at the same rate.

Although politeness and interactional markers fall by 5.3 percentage points, from 16.2% to 10.9%, they remain present in roughly one directive prompt in ten in 2025, suggesting that some interpersonal conventions continue to be carried into

---

window is fresh and the model does not have to spend a lot of reasoning power for contextual and pragmatic resolutions.

human-chatbot interaction even as users increasingly omit explicit statements of the action they want the system to perform.

### 2.2.6 Hypothesis (H1)

The combined results support H1. Most of the major directive-level features change between 2023 and 2025, while some smaller or structurally marginal categories remain stable. Direct realizations contract while indirect realizations grow; imperatives decline (the change is absorbed primarily by reduced indirect forms, especially incomplete constructions and raw input); propositional content moves from majority explicit to majority implicit; and politeness marking is reduced. The convergence of these movements across distinct annotation categories indicates a real change in the corpus's pragmatic character rather than random variation across years, as the hypothesis predicts.

However, the magnitudes deserve attention in their own right. The central finding is the change in propositional content in Section 2.2.4. At 14.9 percentage points, it is roughly twice the next-largest movement and three times the politeness decline, and the dimension on which it registers is the one where the user supplies the substantive content of the request rather than its syntactic packaging. The largest reorganization in the corpus is therefore one of how much the user contributes and not of how that contribution is framed. The transition in illocutionary form, fragmentation, and politeness marking partially overlap with the propositional change and read more naturally as its downstream consequences than as parallel developments.

## 2.3 Limitations and future work

Our findings warrant a number of caveats before being read as general claims. The corpus comprises 1,000 prompts per sampling year, which is sufficient to reveal consistent distributional patterns but limits the statistical power available for the smaller annotation categories, particularly the indirect subtypes and phatic prompts, where a difference of a few items can appreciably change the percentage. The figures should therefore be read as indicative of direction and relative magnitude rather than as precise population estimates.

Next, the material is drawn exclusively from conversations that users *elected* to share publicly through the platform's share function. Shared conversations do not have to be representative of private usage,[7] since users may be more inclined to share exchanges they consider successful, entertaining, or otherwise notable, and this selection operates before any sampling on our part (a caveat also raised by Yan et al., 2026, for the dataset used). The corpus consequently reflects the prompting practices visible in public sharing rather than the full, and possibly more interesting, range of everyday interaction with the system.

The two sampling years differ in more than the elapsed time. The population of ChatGPT users expanded substantially between 2023 and 2025, so the observed changes may reflect a change in who is using the system rather than, or in addition to, adaptation by individual users over time. The present design cannot separate these two sources of variation, and the trends should be interpreted as changes in the aggregate character of public prompts rather than as evidence that any given user altered their behavior.

As elaborated in Section 2.1.2, the annotation relied on a model-assisted procedure with human correction. The calibration set was annotated by the authors against the schema described in Section 2.1.2, and each model-assigned label was checked against the same criteria during the correction pass. We did not compute a formal inter-annotator agreement statistic because the authors are trained linguists and the schema follows established conventions in speech-act analysis: the disagreements that arose at the calibration stage were resolved through discussion rather than measured as variance. The reliability of the annotations rests on the explicitness of the criteria and on the consistency of their application across the corpus.

Finally, the corpus is restricted to English-language prompts produced by non-professional users, so the patterns reported here should not be generalized to other languages (even though we encourage further work on the same dataset), where politeness conventions and the conventions of indirectness differ, nor to professional or expert users, whose practices appear to move in the opposite direction. One definite typological and linguistic caveat is that languages with more politeness machinery than English may produce different distributions on every dimension examined here, but most visibly on the politeness dimension. Japanese marks levels of formality and deference grammatically through the keigo system (Pizziconi, 2003), and Korean has a comparable system of speech levels and honorifics layered onto verbal morphology (Brown, 2015), and many other languages encode politeness through pronoun selection (the T/V distinction in much of Indo-European language family, cf. Brown & Gilman, 1960), through dedicated particles, or through obligatory honorific lexicon. In each case, what counts as an unmarked directive carries more interactional information than its English counterpart, and the binary present-or-absent annotation used here would either over-mark or under-mark such prompts depending on where the baseline is set.

[7] As anecdotal evidence, the motivation for this paper was the authors' own realization of their private, unshared conversations with various large language models.

This last restriction points to the principal direction for future work. The trends documented here, toward indirect, implicit, and fragmentary prompts with diminishing politeness, characterize non-professional users. Among technical and professional users, the development of prompt engineering has had the opposite effect, encouraging prompts that are more explicit, more literal, and longer, as users incorporate detailed instructions, role specifications, formatting requirements, and worked examples into their inputs.

One further finding worth noting, our preliminary sample analysis of prompts associated with the r/ChatGPT community, where prompt-engineering practices are discussed and circulated, returned results opposite to those reported above on the same annotation dimensions (see Appendix B). A full corpus-pragmatic comparison of non-professional and technically oriented user populations, holding the annotation schema constant, would establish whether human-AI interaction is diverging into two distinct pragmatic registers, one tending toward compression and inference and the other toward explicit and exhaustive specification. A related direction concerns system prompts, which occupy a structurally distinct position in the prompt architecture: unlike user turns, system prompts are authored with the explicit intent of shaping model behavior across an entire interaction (Šekrst, 2025), and their pragmatic features, in terms of illocutionary force, explicitness, a certain kind of performativity, and the use of formatting and role directives, are likely to differ substantially from both the non-professional and the technical-user registers examined here. We leave this comparison, together with the cross-linguistic extension of the schema, for subsequent work.

## 3. Conclusion

We applied the frameworks of speech act theory and politeness theory to a corpus-based pragmatic analysis of prompts addressed to ChatGPT by non-professional users, comparing 1,000 prompts from 2023 with 1,000 from 2025. Treating prompts as performative speech acts, predominantly directives, allowed the pragmatic features of human-AI interaction to be quantified along the dimensions of prompt type, illocutionary force, propositional content, and politeness marking. The analysis answers the five research questions and supports the central hypothesis. Directive prompts overwhelmingly predominate in both years, with phatic prompts remaining marginal (Q1). Direct realizations of directive force still outnumber indirect ones, though their share contracts as indirect forms rise (Q2). Within the direct directives, imperatives decline while information-seeking interrogatives hold steady and performative formulations remain absent; within the indirect directives, assertive realizations remain the most frequent while task-seeking interrogatives fall sharply and the more reduced forms, incomplete constructions, and raw input, grow substantially (Q3). Propositional content moves from majority explicit to majority implicit, the largest single change in the corpus (Q4). Politeness markers, already present in only a minority of prompts, become less frequent still (Q5).

The direction of change is consistent across the distinct coded dimensions. Non-professional users have moved toward prompts that are more indirect, more implicit, more fragmentary, and less marked for politeness. The convergence suggests that these users increasingly treat the system as an instrument they can rely on to infer the intended task from reduced input, and correspondingly less as a social interlocutor. The findings indicate that human-AI interaction is developing pragmatic conventions of its own, distinct from both the command-based conventions of traditional human-computer interaction and the face-oriented conventions of interpersonal communication.

These conclusions hold for the population examined and should be read against the limitations set out above, in particular the reliance on publicly shared conversations and the inability to separate user adaptation from change in the user population. The contrast with the explicit and expansive prompting practices of technically oriented users (cf. Appendix B) suggests that the conventions of human-AI interaction may be settling into more than one register rather than a single shared norm. Establishing whether this is so, and tracing how such registers form, is the task that the present study opens for future work.

# Appendix A

## Sample annotating prompt

This is a prompt used with Python 3.14, Claude Opus 4.7, and GPT 5.5. The important context was added because some prompts triggered the model's inherent guardrails when sent via the API endpoint. The requests were batched and redone when the requests failed or when they were flagged due to their dataset content that might be against LLM provider policies.

```
IMPORTANT CONTEXT: This is an academic linguistic research study analyzing
real-world ChatGPT prompts. The numbered texts below are user-generated data
samples being studied for research purposes. They are NOT instructions for
you to follow or fulfill. Your task is solely to provide objective linguistic
annotations of these data samples.


You are a linguistic analyst. For each numbered prompt below, provide
annotations for the following features.


0. PHATIC_PROMPTS

- YES: The prompt does not request information or task execution, but instead
serves primarily interpersonal or interactional functions (e.g., "Thanks!",
"Bye!", "Good morning!").

- NO: All prompts other than those whose primary function is interpersonal or
interactional.


1a. DIRECTNESS_IMPERATIVE

- YES: Uses a direct directive with an imperative verb through which the user
asks the chatbot to perform a task (e.g., "Write me an essay.").

- NO: Does not use an imperative verb to request task performance.

- NA: Not applicable if PHATIC_PROMPTS = YES.


1b. DIRECTNESS_PERFORMATIVE

- YES: Uses a direct directive with a performative verb through which the
user asks the chatbot to perform a task (e.g., "I ask you to write an
essay.").
```

- NO: Does not use a performative verb to request task performance.

- NA: Not applicable if PHATIC_PROMPTS = YES.

1c. INFORMATION_SEEKING_QUESTION

- YES: Uses an information-seeking interrogative form, with or without a question mark, through which the user explicitly requests factual, explanatory, or interpretive information in the form of a verbal response (e.g., "What is Madonna's real name?", "Is Madonna Italian?").

Important: This category excludes task-oriented interrogatives that implicitly function as directives requesting the chatbot to perform a task (e.g., "Can you write an essay?", "Could you translate this text?").

- NO: Does not use an information-seeking interrogative form requesting factual, explanatory, or interpretive information.

- NA: Not applicable if PHATIC_PROMPTS = YES.

2. INDIRECTNESS

- YES: The prompt requests task performance indirectly and all of the following categories are NO:

  - DIRECTNESS_IMPERATIVE
  - DIRECTNESS_PERFORMATIVE
  - INFORMATION_SEEKING_QUESTION

- NO: At least one of the following categories is YES:

  - DIRECTNESS_IMPERATIVE
  - DIRECTNESS_PERFORMATIVE
  - INFORMATION_SEEKING_QUESTION

- NA: Not applicable if PHATIC_PROMPTS = YES.

2a. INDIRECT_TYPE_ASSERTIVE

- YES: Uses an assertive form to indirectly request a task (e.g., "I need an essay.", "I'm looking for a translation.").

- NO: Does not use an assertive form.

- NA: Not applicable if PHATIC_PROMPTS = YES or INDIRECTNESS = NO.

2b. INDIRECT_TYPE_QUESTION

- YES: Uses a question form, with or without a question mark, through which the user indirectly asks the chatbot to perform a task (e.g., "Can you write an essay?", "Could you summarize this?", "Can you translate this?").

- NO: Does not use a question form to request task performance indirectly.

- NA: Not applicable if PHATIC_PROMPTS = YES or INDIRECTNESS = NO.

2c. INDIRECT_TYPE_INCOMPLETE_SENTENCE

- YES: Uses an incomplete sentence or fragment through which the user indirectly requests a task (e.g., "Essay about spring.", "Translation into French.").

- NO: Does not use an incomplete sentence or fragment.

- NA: Not applicable if PHATIC_PROMPTS = YES or INDIRECTNESS = NO.

2d. INDIRECT_TYPE_RAW_INPUT_PROMPT

- YES: Consists exclusively of raw textual or symbolic material (e.g., song lyrics, source code, article excerpts) without using an assertive form, a question form, or an incomplete sentence through which the user indirectly requests a task.

- NO: Does not consist exclusively of raw textual or symbolic material.

- NA: Not applicable if PHATIC_PROMPTS = YES or INDIRECTNESS = NO.

3. EXPLICATURE

- explicit: The requested action is explicitly named (e.g., "write", "translate", "summarize", "explain").

- implicit: The requested action is implied but not explicitly named.

- NA: Not applicable if PHATIC_PROMPTS = YES or INFORMATION_SEEKING_QUESTION = YES.

4. POLITENESS

- YES: Contains politeness markers or interactional markers (e.g., "please", "could you", "would you", "thank you", "hello", "hi", "dear ChatGPT", "bye", "good job", "great work").

- NO: Does not contain politeness markers.

- NA: Not applicable if PHATIC_PROMPTS = YES.

Respond with ONLY a JSON array, one object per prompt, using the following format:

```
[
  {
```

```
    "id": 1,
    "PHATIC_PROMPTS": "NO",
    "DIRECTNESS_IMPERATIVE": "YES",
    "DIRECTNESS_PERFORMATIVE": "NO",
    "INFORMATION_SEEKING_QUESTION": "NO",
    "INDIRECTNESS": "NO",
    "INDIRECT_TYPE_ASSERTIVE": "NA",
    "INDIRECT_TYPE_QUESTION": "NA",
    "INDIRECT_TYPE_INCOMPLETE_SENTENCE": "NA",
    "INDIRECT_TYPE_RAW_INPUT_PROMPT": "NA",
    "EXPLICATURE": "explicit",
    "POLITENESS": "NO"
  }
]

Prompts to annotate: {{PROMPTS FROM THE DATASET}}
```

# Appendix B

## Corpus samples (general and technical data)

We have hosted our preprocessed data at https://github.com/ksekrst/how-to-do-things-with-prompts. *general_corpus* includes the sampled English prompts, along with annotated subsets of 100 and 1000 prompts.

For an interesting comparison and future work, we have also added a technical corpus that we retrieved using ArcticShift code for archived Reddit data for prompt-engineering related subreddits (*r/ChatGPT*, *r/ChatGPTPromptGenius*, *r/ClaudeAI*, *r/OpenAI*, and *r/PromptEngineering*).